\documentclass[sigconf]{acmart}
\renewcommand\footnotetextcopyrightpermission[1]{}

\usepackage{amsmath,amsfonts}
\usepackage{graphicx}
\usepackage{xcolor}
\usepackage{xspace}
\usepackage{listings}
\usepackage{booktabs}
\usepackage{makecell}
\usepackage{tabularx}
\usepackage{multirow}
\usepackage{colortbl}
\usepackage{tikz}
\usepackage{nicematrix}
\usepackage{pifont}
\usepackage{float}
\usepackage{balance}
\usepackage{algorithm}
\usepackage{algpseudocode}
\usepackage{subcaption}
\usepackage{makecell}
\begin{document}

\newcommand{\name}{LLM-FSM\xspace}

\newcommand{\datasetsize}{1,000\xspace}

\newcommand{\humansize}{100\xspace}

\newcommand{\cmark}{\textcolor{green!60!black}{\ding{51}}} 
\newcommand{\xmark}{\textcolor{red!90!black}{\ding{55}}}   

\newcommand{\yuheng}[1]{\textcolor{green}{\textbf{Yuheng:} #1}}

\newcommand{\berk}[1]{\textcolor{blue}{\textbf{Berk:} #1}}

\newcommand{\zhouhua}[1]{\textcolor{orange}{\textbf{Zhouhua:} #1}}

\newcommand{\peijing}[1]{\textcolor{red}{\textbf{Peijing:} #1}}

\newcommand{\caroline}[1]{\textcolor{purple}{\textbf{Caroline:} #1}}

\newcommand{\priyanka}[1]{\textcolor{green}{\textbf{Priyanka:} #1}}

\newcommand{\thierry}[1]{\textcolor{red}{\textbf{Thierry:} #1}}

\author{Yuheng Wu}
\affiliation{
  \institution{Stanford University}
  \city{Stanford}
  \state{CA}
  \country{USA}
}
\email{yuhengwu@stanford.edu}

\author{Berk Gokmen}
\affiliation{
  \institution{Stanford University}
  \city{Stanford}
  \state{CA}
  \country{USA}
}
\email{bgokmen@stanford.edu}

\author{Zhouhua Xie}
\affiliation{
  \institution{Stanford University}
  \city{Stanford}
  \state{CA}
  \country{USA}
}
\email{xzh015@stanford.edu}

\author{Peijing Li}
\affiliation{
  \institution{Stanford University}
  \city{Stanford}
  \state{CA}
  \country{USA}
}
\email{peli@stanford.edu}

\author{Caroline Trippel}
\affiliation{
  \institution{Stanford University}
  \city{Stanford}
  \state{CA}
  \country{USA}
}
\email{trippel@stanford.edu}

\author{Priyanka Raina}
\affiliation{
  \institution{Stanford University}
  \city{Stanford}
  \state{CA}
  \country{USA}
}
\email{praina@stanford.edu}

\author{Thierry Tambe}
\affiliation{
  \institution{Stanford University}
  \city{Stanford}
  \state{CA}
  \country{USA}
}
\email{ttambe@stanford.edu}

\title{LLM-FSM: Scaling Large Language Models for Finite-State Reasoning in RTL Code Generation}

\begin{abstract}

Finite-state reasoning, the ability to understand and implement state-dependent behavior, is central to hardware design. 
In this paper, we present \name, a benchmark that evaluates how well large language models (LLMs) can recover finite-state machine (FSM) behavior from natural-language specifications and translate it into correct register transfer-level (RTL) implementations. 
Unlike prior specification-to-RTL benchmarks that rely on manually constructed examples, \name is built through a fully automated pipeline.
\name first constructs FSM with configurable state counts and constrained transition structures. It then prompts LLMs to express each FSM in a structured YAML format with an application context, and to further convert that YAML into a natural-language (NL) specification.
From the same YAML, our pipeline synthesizes the reference RTL and testbench in a correct-by-construction manner. 
All \datasetsize problems are verified using LLM-based and SAT-solver-based checks, with human review on a subset.
Our experiments show that even the strongest LLMs exhibit sharply declining accuracy as FSM complexity increases. 
We further demonstrate that training-time scaling via supervised fine-tuning (SFT) generalizes effectively to out-of-distribution (OOD) tasks, while increasing test-time compute improves reasoning reliability. Finally, \name remains extensible by allowing its FSM complexity to scale with future model capabilities.
\end{abstract}

\maketitle
\fancyhead{}

\section{Introduction}\label{sec:intro}

Large language models (LLMs) have demonstrated strong reasoning capabilities in tasks such as math problem solving and code generation \cite{guo2025deepseek}. Recently, there has been growing interest in applying LLMs to assist electronic design automation (EDA) tasks \cite{thakur_2023_benchmarking, chang_2023_chipgpt}. A representative example is register transfer-level (RTL) code generation, where a model receives a 
natural-language (NL) design specification and generates the corresponding implementation \cite{lu_2024_rtllm,liu_2023_verilogeval,liu_2024_openllmrtl, pinckney_2025_revisiting}. NL is the default interface in LLM-based design workflows, since users describe intended behavior in NL rather than formal specification formats. To improve the performance of LLMs on such NL specification-to-RTL tasks, researchers have explored techniques such as supervised fine-tuning (SFT) \cite{liu_2025_craftrtl}, reinforcement learning (RL) \cite{wang_2025_verireason}, and multi-agent collaboration \cite{zhao_2025_mage}. All these approaches can be viewed through the lens of \textit{scaling}, which broadly refers to allocating more computation, during training or inference, to enhance a model's RTL reasoning performance.

In evaluating LLMs' ability to generate RTL, an important component is their finite-state reasoning capability, which refers to the ability to understand and implement state-dependent behavior. This capability underlies a wide range of hardware workflows, including controllers, protocols, and multi-cycle sequencing logic. However, there is no benchmark designed specifically to evaluate LLMs on finite-state reasoning. Existing specification-to-RTL datasets \cite{liu_2023_verilogeval, lu_2024_rtllm, liu_2024_openllmrtl, pinckney_2025_revisiting, pinckney_2025_comprehensive,purini_2025_archxbench} contain only a small number of such tasks, and each example requires domain experts to manually write the specification, reference RTL, and testbench, making them difficult to scale. Therefore, our research question is: 
\textit{How can we automatically create a large and scalable NL specification-to-RTL benchmark that evaluates LLMs on finite-state reasoning?}

In this paper, we present \name, a large-scale NL specification-to-RTL benchmark designed to evaluate finite-state reasoning in LLMs. Unlike existing benchmarks that rely on manual construction, our approach fully automates the dataset curation process. The resulting benchmark is both scalable and controllable, enabling new problems to be generated at configurable levels of finite-state machine (FSM) complexity. All reference RTL implementations and testbenches are produced in a correct-by-construction manner. The dataset contains problems spanning diverse and realistic finite-state reasoning scenarios, providing a challenging and extensible benchmark for evaluating LLM-based RTL generation.

\begin{figure*}[t]
    \centering
    \includegraphics[width=\linewidth]{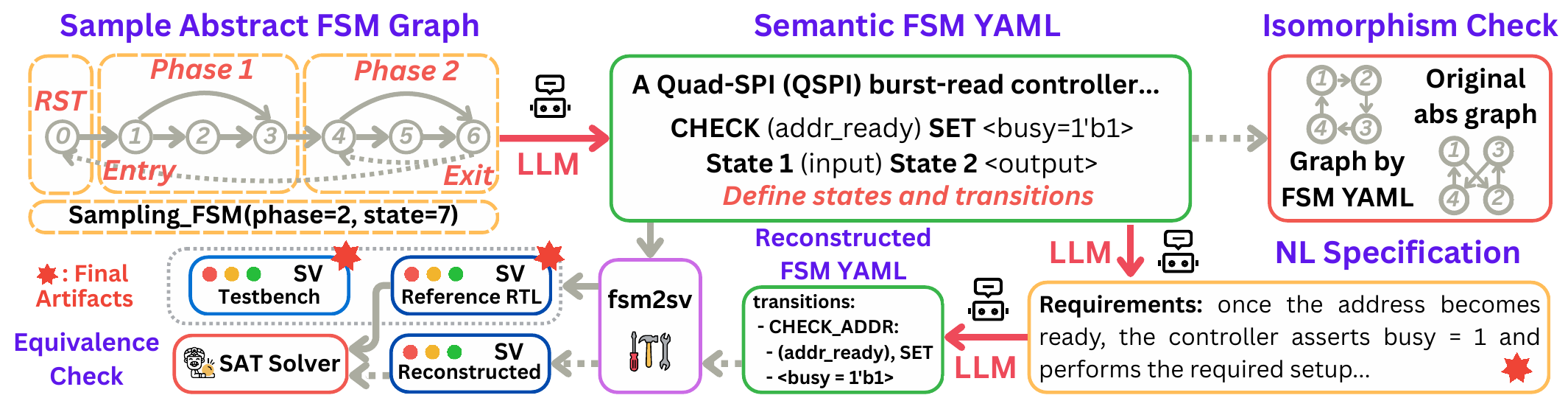}
    \caption{Overview of the \name data curation pipeline.
    The process begins by constructing an abstract FSM graph, followed by LLM-based specification generation, automatic RTL and testbench synthesis, and isomorphism/equivalence check.}
    \label{fig:pipeline} \vspace{-0.5em}
\end{figure*}

\begin{figure*}[t]
    \centering
    \includegraphics[width=\linewidth]{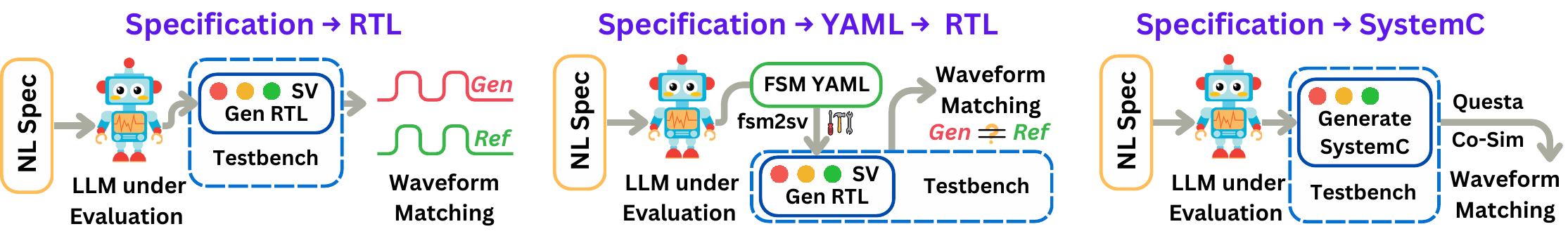}
    \caption{Overview of the \name evaluation pipeline. An NL specification is processed through three tool-chain settings: Specification$\rightarrow$RTL, Specification$\rightarrow$YAML$\rightarrow$RTL, and Specification$\rightarrow$SystemC. Each model prediction is executed under the same reference testbench, and correctness is determined by cycle-by-cycle output matching against the reference RTL.} \vspace{-0.5em}
    \label{fig:evaluation-pipeline}
\end{figure*}

As illustrated in Figure~\ref{fig:pipeline}, the data curation process begins by sampling an \textbf{abstract FSM graph} based solely on the specified number of states and a set of structural constraints. These constraints ensure that all states are reachable from reset, transitions form a connected directed graph, and the out-degree and back-edge probabilities lie within prescribed bounds. This graph encodes only the topology of the machine and carries no application-level meaning. An LLM then interprets this abstract graph and generates an application context consistent with its transition structure, assigning \textbf{semantic} roles to each state and defining the inputs, outputs, and conditions associated with each transition. This FSM description is stored in a \textbf{structured FSM YAML format}. We check that the FSM YAML representation is isomorphic to the original abstract graph and discard any samples that fail this check. Using the fsm2sv \cite{fsm2sv} tool, we translate the full FSM YAML into a SystemVerilog implementation. We additionally design a generator that constructs a testbench by systematically traversing the transitions encoded in the FSM YAML. Together, these steps yield the \textbf{reference RTL implementation} and its corresponding \textbf{testbench}.

In the next stage, we use an LLM to translate the FSM YAML into an \textbf{NL specification} that describes the behavior of the machine. Given the state mappings contained in the YAML, the LLM is then asked to reconstruct the FSM by converting the specification back into YAML format. Using the fsm2sv tool once again, we generate SystemVerilog code from both the \textbf{reconstructed FSM} and the original FSM. The two designs are then passed to Yosys~\cite{yosys-eqy} for SAT-solver-based equivalence checking to determine whether the specification preserves the behavior of the original FSM. Specifications that fail the equivalence check are discarded, and a subset of the
passing specifications undergoes human review to verify narrative
clarity and hardware plausibility of the NL description.

Our curated \name dataset contains \datasetsize problems, grouped into three
difficulty tiers (easy, medium, and hard) based on state number and edge complexity. As shown in Figure~\ref{fig:evaluation-pipeline}, we evaluate models under three pipelines: (1) \textbf{Specification$\rightarrow$RTL}, which tests end-to-end Verilog generation; (2) \textbf{Specification$\rightarrow$YAML$\rightarrow$RTL}, where the model constructs the FSM in a YAML format and RTL is generated using fsm2sv; and (3) \textbf{Specification$\rightarrow$SystemC}, where the model produces SystemC for validation through SystemC-Verilog co-simulation (Co-Sim). Across these settings, we evaluate a wide range of frontier models and find that even the most advanced LLMs struggle on the hard tier of \name.

Our results show that \name exposes clear limitations of current LLMs in finite-state reasoning and offers a challenging benchmark for evaluating future models. We further demonstrate that both training-time and test-time scaling yield consistent performance improvements; specifically, SFT on our data boosts model generalization across both in-distribution (ID) and out-of-distribution (OOD) tasks. In summary, our work makes the following contributions:
\begin{itemize}
    \item An automatic pipeline that synthesizes scalable RTL benchmarks, verified through a SAT solver, and designed to evolve jointly with advances in LLM capabilities;
    \item A rigorous evaluation across three tool-chain pipelines, demonstrating that existing models struggle on \name and underscoring the benchmark's challenging nature;
    \item An analysis of scaling dynamics, demonstrating that training-time scaling via SFT generalizes effectively to OOD tasks, while parallel test-time sampling outperforms serial decoding for finite-state reasoning.
\end{itemize}

\section{Related Work}\label{sec:related}
In this section, we review related work in three areas: the application of LLMs to RTL code generation, existing benchmarks for evaluating RTL code generation tasks, and recent advances in scaling LLMs to enhance reasoning capabilities.

\begin{table}[t] 
\centering
\small
\caption{Comparison of RTL code generation benchmarks. The five dimensions are: (1) \textit{Dataset Size}: the number of test instances included in the benchmark; (2) \textit{Automated Generation}: the dataset is automatically constructed and scalable; (3) \textit{Difficulty Control}: problem difficulty can be adjusted; (4) \textit{Realistic Specification}: tasks are described in NL specifications; and (5) \textit{Automated Verification}: testbenches are automatically generated for validation.}
\resizebox{\columnwidth}{!}{
\begin{tabular}{@{}lccccc@{}} 
\toprule
\textbf{Benchmark} & \textbf{\makecell{Dataset \\ Size}} & \textbf{\makecell{Automated \\ Generation}} & \textbf{\makecell{Difficulty \\ Control}} & \textbf{\makecell{Realistic \\ Specification}} & \textbf{\makecell{Automated \\ Verification}}\\
\midrule
RTLLM v1 \cite{lu_2024_rtllm} & 29 & \xmark & \xmark & \cmark & \xmark \\
RTLLM v2 \cite{liu_2024_openllmrtl} & 50 & \xmark & \xmark & \cmark & \xmark \\
VerilogEval v1 \cite{liu_2023_verilogeval} & 156& \xmark & \xmark & \cmark & \xmark \\
VerilogEval v2 \cite{pinckney_2025_revisiting} & 156& \xmark & \xmark & \cmark & \xmark \\
ArchXBench \cite{purini_2025_archxbench} & 51 & \xmark & \cmark & \cmark & \xmark \\
CVDP \cite{pinckney_2025_comprehensive} & 783 & \xmark & \cmark & \cmark & \xmark \\
\textbf{LLM-FSM (Ours)} & 1000 & \cmark & \cmark & \cmark & \cmark 
 \\
\bottomrule
\end{tabular}%
}

\label{tab:compare}
\end{table}

\subsection{LLMs for RTL Code Generation}
Researchers have explored using LLMs for hardware code generation since the rise of LLMs \cite{thakur_2023_benchmarking, chang_2023_chipgpt}. However, due to the scarcity of hardware description language (HDL) data compared to high-level languages such as Python or C++ \cite{tasnia_2025_opl4gpt}, pretrained models often perform poorly on hardware-related tasks \cite{liu_2024_rtlcoder}. To address this limitation, recent studies have focused on SFT of LLMs on domain-specific corpora to better adapt them for RTL design and synthesis \cite{liu_2024_rtlcoder, thakur_2024_verigen, pei_2024_betterv, liu_2025_craftrtl, cui_2024_origen, wei_2025_vericoder, akyash_2025_rtl,deng_2025_scalertl, calzada_2025_verilogdb}. 
Beyond SFT, some works further employ RL-based post-training \cite{guo2025deepseek, wang_2025_large, wang_2025_verireason} to enhance code correctness.
At the inference stage, many approaches design workflows incorporate execution feedback \cite{tasnia_2025_veriopt} and leverage multi-agent collaboration \cite{bardianadimi_2024_a, zhao_2025_mage, yu_2025_spec2rtlagent} to further enhance the quality and reliability of generated RTL code.

Beyond direct RTL code generation, LLMs have also been explored for higher-level hardware design workflows \cite{ravindran_2025_survey}. Some approaches first generate high-level languages such as C or Python, which are then translated into domain-specific representations \cite{liao_2024_are, batten_2024_pyhdleval}. 
In addition, LLMs are increasingly applied to other stages of the hardware design flow, including testbench generation \cite{ma_2024_verilogreader, bhandari_2024_llmaided, qiu_2024_autobench, qiu_2025_correctbench}, formal specification synthesis \cite{sun_2023_towards, yan_2025_assertllm}, and temporal logic specification generation \cite{cosler_2023_nl2spec, mendoza_2024_translating}.

\subsection{RTL Code Generation Benchmarks}
As shown in Table~\ref{tab:compare}, several benchmarks have been developed to evaluate LLM performance on RTL code generation. Among them, the two most widely used are RTLLM \cite{lu_2024_rtllm, liu_2024_openllmrtl} and VerilogEval \cite{liu_2023_verilogeval, pinckney_2025_revisiting}. RTLLM contains 50 problems, while VerilogEval includes 156 tasks covering both combinational logic modules and FSMs. ArchXBench \cite{purini_2025_archxbench}, on the other hand, consists of 51 human-authored high-level RTL design tasks, such as generating FFT and CNN modules. In all these benchmarks, the specifications and corresponding testbenches are manually written, making them difficult to scale to larger datasets.
In addition, RTL-Repo \cite{allam_2024_rtlrepo} provides a fill-in-the-blank style task focusing on completing partial RTL code, and CVDP \cite{pinckney_2025_comprehensive} offers 783 human-written questions covering the broader RTL design pipeline. Although larger in scale, these datasets are still manually curated and span diverse tasks such as debugging, code comprehension, and design analysis, rather than focusing on specification-to-RTL generation.

\subsection{Scaling LLMs for Reasoning}
Scaling refers to allocating more computation to enhance the reasoning capabilities of LLMs. Such scaling can occur both during training and at inference time. At the training stage, scaling can be achieved through RL \cite{guo2025deepseek} or by SFT on reasoning traces distilled from stronger teacher models \cite{li_2025_llms}. These approaches encourage the model to generate longer and more coherent reasoning chains, resulting in improved performance on complex reasoning tasks.

At the inference stage, scaling is commonly referred to as test-time scaling (TTS). It can be realized by encouraging deeper and more deliberate reasoning within a single inference path \cite{wei2022chain, muennighoff_2025_s1}. Alternatively, multi-trace TTS \cite{brown_2024_large, snell2025scaling, wu2025tom, wu2025on} samples multiple candidate completions in parallel and selects the best one using either verifier-based evaluation \cite{wang2024math} or voting-based aggregation \cite{wangself, wang_2025_ranked}. Recent studies further integrate search algorithms \cite{yao2023tree, biforest} that interleave generation and selection in a step-by-step manner, refining the output through iterative exploration and verification.

\begin{figure*}[t]
    \centering
    \includegraphics[width=\linewidth]{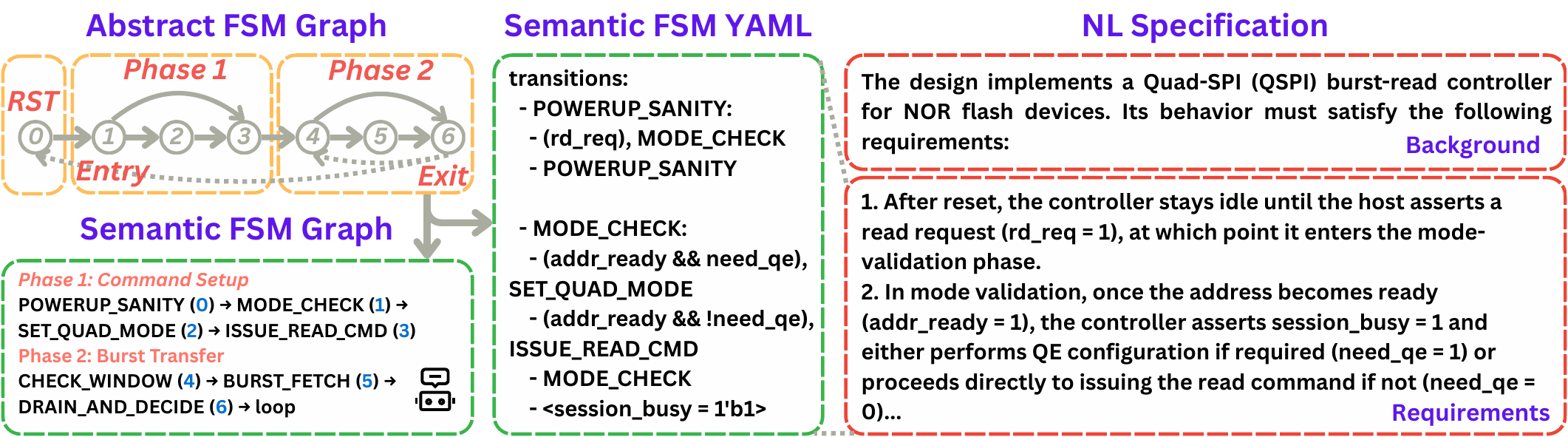}
    \caption{An example illustrating the generation process.
The abstract graph is first sampled topologically, and an LLM then assigns semantics, here producing a Quad-SPI burst-read controller for a NOR-flash device.}
    \label{fig:example} \vspace{-0.5em}
\end{figure*}

\section{Methods}\label{sec:method}

In this section, we describe how \name is constructed and how models are evaluated on it. 
We first synthesize abstract FSM topologies (Section~\ref{sec:abstract-graph}), enrich them with LLM-generated semantics (Section~\ref{sec:semantic-fsm}), and compile them into RTL and testbenches (Section~\ref{sec:rtl-tb-gen}), validating consistency through SAT-based equivalence (Section~\ref{sec:spec-consistency}). 
We then present dataset statistics, an analysis of alignment with real-world applications, human validation, and the evaluation pipelines used to assess model performance (Sections~\ref{sec:curation-dynamics}--\ref{sec:evaluation-pipeline}).

\subsection{Abstract FSM Graph Construction}
\label{sec:abstract-graph}

\paragraph{Phase-based abstract graph structure.}
We represent each FSM using a two-level structure organized into phases. A phase corresponds to a coherent stage of operation, such as initialization, data transfer, or error handling. Each phase is a subgraph with a single entry and exit, and all internal states lie on paths between them. Phase transitions are modeled by directed edges from the exit of one phase to the entry of another, allowing the abstract graph to capture high-level control flow.

\paragraph{Topology generation algorithm.}
For each phase, we generate a minimal chain from entry to exit to ensure reachability, then add forward branches, back edges, and self-loops under user-controlled probabilities while capping the out degree. 
We add a reset block and connect phases in a simple cycle to guarantee global reachability. This ensures that every sampled abstract FSM is structurally valid.
Additional inter-phase edges are sampled to create jumps between phases. This procedure produces an abstract FSM graph whose structure is determined by a small set of topology parameters.

\paragraph{Example}
Figure~\ref{fig:example} shows an example with two phases. Each phase contains an entry–exit chain with additional forward branches, back edges, and self-loops sampled under the specified probabilities, while graph-level edges form a cycle that guarantees global reachability. This two-level organization is intentional: real hardware controllers typically consist of several semantically coherent phases assembled into a larger control graph. Directly sampling an unconstrained flat FSM makes it difficult for an LLM to assign meaningful roles to states or to construct realistic hardware scenarios. By generating topology at both the phase level and the graph level, we obtain abstract FSMs that are structurally rich yet still amenable to consistent semantic interpretation.

\subsection{Semantic FSM Generation and YAML Construction}
\label{sec:semantic-fsm}

\paragraph{LLM-based semantic FSM generation.}
As shown in Figure~\ref{fig:example}, given an abstract FSM graph from Section~\ref{sec:abstract-graph}, we use an LLM to turn this purely structural object into a semantic FSM. The prompt exposes the phase structure, the exact edge list, and asks the model to choose a realistic hardware scenario, assign descriptive names to all states, and design input and output signals. The model then produces an fsm2sv-compatible semantic FSM YAML file that specifies reset behavior, input and output declarations, and for each state, a list of guarded transitions and outputs that follow the provided connectivity. In addition to the YAML, the model also generates a short story of the workflow.

\paragraph{Graph isomorphism verification.}
As shown in Figure~\ref{fig:pipeline}, let $G_{\text{abs}} = (V, E)$ denote the abstract FSM graph and let $G_{\text{yaml}} = (\hat V, \hat E)$ be the graph obtained from the transitions in the generated YAML. The state mapping produced by the LLM defines a candidate bijection $f \colon V \to \hat V$, and we require
\[
(u, v) \in E \quad \Longleftrightarrow \quad (f(u), f(v)) \in \hat E
\]
for all $u, v \in V$. Any YAML instance that violates this isomorphism condition is discarded, so the retained semantic FSMs add meaning while preserving the original topology.

\subsection{Reference RTL and Testbench Generation}
\label{sec:rtl-tb-gen}

\paragraph{Reference RTL}
Given a checked semantic FSM YAML, the reference implementation is
generated automatically using the fsm2sv tool, as shown in Figure~\ref{fig:pipeline}. Because the
YAML format fully specifies each state, its outputs, and the ordered
conditional transitions, the translation to synthesizable SystemVerilog is
mechanical: every YAML transition becomes a guarded branch in the
\texttt{always\_comb} block, and state encodings are assigned in a consistent
one-hot or counter style. Since the YAML itself has already passed the
topology-preserving isomorphism check, the resulting RTL is
correct-by-construction relative to the input FSM.

\paragraph{Testbench generation.}
The fsm2sv package does not include a testbench generator, so we extend
the tool with a testbench synthesis module. The key requirement is to produce a
set of input sequences that covers all states and all transitions of the FSM.
Let the FSM be a directed graph $G = (V,E)$ with initial state $s_0$. For every
edge $e = (u,v) \in E$, the generator performs:
\[
\text{find a path } s_0 \leadsto u,\qquad
\text{emit inputs that satisfy the guard of } e.
\]
This guarantees that each transition is exercised at least once. The generator runs in polynomial time in the size of the FSM. It also emits additional random stimuli to improve robustness. The resulting testbench is a self-contained module that instantiates the DUT, drives the generated input sequences cycle by cycle, and records waveforms for debugging.

\subsection{Specification Generation and Formal Verification}
\label{sec:spec-consistency}
\paragraph{NL specification synthesis.}
As shown in Figure~\ref{fig:pipeline}, given a topology-verified YAML FSM \(\mathcal{Y}\), we first ask an LLM to produce an NL specification \(\Sigma = (\Sigma_{\text{IO}}, \Sigma_{\text{req}})\). The prompt exposes only the inputs/outputs, reset configuration, and transition table, and requires: (1) an Inputs and Outputs section \((\Sigma_{\text{IO}})\) that lists every signal using the exact YAML names; and (2) a Requirements section \((\Sigma_{\text{req}})\) that paraphrases each group of transitions into requirements. This defines a forward map
\(
F:\mathcal{Y}\rightarrow\Sigma
\)
that hides state names but keeps the transition semantics.

To check that the specification is semantically complete, we perform a second LLM pass that reconstructs a fsm2sv-compatible YAML \(\widetilde{\mathcal{Y}}\) from \(\Sigma\) and the state mapping. Invalid or structurally inconsistent reconstructions are discarded, leaving only pairs \((\mathcal{Y}, \widetilde{\mathcal{Y}})\) where both directions
\(
\mathcal{Y}\xrightarrow{F}\Sigma\xrightarrow{G}\widetilde{\mathcal{Y}}
\)
succeed.

\paragraph{SAT-based equivalence checking.}

For each structurally correct pair \((\mathcal{Y}, \widetilde{\mathcal{Y}})\), we compile both YAML files with fsm2sv to produce two RTL machines with identical I/O interfaces:
\[
  \mathcal{M}=(S,s_0,I,O,\delta,\lambda),\qquad
  \widetilde{\mathcal{M}}=(\widetilde S,\widetilde s_0,I,O,
  \widetilde\delta,\widetilde\lambda).
\]

To check whether the NL specification preserves the behavior of the original FSM, we use Yosys's equivalence-checking flow
(\texttt{equiv\_make}, \texttt{equiv\_simple}, \texttt{equiv\_struct}, \texttt{equiv\_status}).  
This constructs a sequential miter between the two RTL designs and searches for an input sequence under which their outputs diverge.

Formally, let the two machines process the same input sequence \((x_0,\dots,x_T)\).  
A mismatch at time \(t\) is recorded as
\[
d_t = \bigl(\lambda(s_t,x_t)\neq\widetilde{\lambda}(\widetilde{s}_t,x_t)\bigr),
\qquad
D = \bigvee_{t\le T} d_t .
\]

The solver asks whether some reachable execution can produce \(D=1\).  
If a counterexample trace is found, the sample is discarded.  
If Yosys completes the check without reporting a mismatch, we accept the pair as behaviorally equivalent for all executions explored by the checker.
Only examples that pass this equivalence check are kept.  
This round-trip filter ensures that the NL specification is consistent with the transition semantics encoded in the original YAML FSM.

\begin{table}[t]
\caption{\textbf{\name dataset statistics.}
Tasks are grouped into three difficulty tiers based on total state count.
For each tier, we report dataset size, complexity measures, specification word count, and reference RTL code lines.}
\label{tab:dataset-stats}
\centering
\resizebox{\linewidth}{!}{
\begin{tabular}{lcccccc}
\toprule
\textbf{Tier} & \textbf{Count} & \textbf{States} &
\textbf{\makecell{Avg.\\Edges}} &
\textbf{\makecell{Avg.\\Phases}} &
\textbf{\makecell{Avg. Spec.\\Word Count}} &
\textbf{\makecell{Avg. Ref.\\Code Lines}} \\
\midrule
Low    & 334 & 4--14  & 11.95 & 2.71 & 409.3  & 154.8 \\
Medium & 333 & 14--27 & 32.17 & 5.24 & 780.3 & 301.8 \\
High   & 333 & 27--59 & 65.39 & 8.83 & 1265.1 & 501.3 \\
\midrule
\textbf{Overall} & 1000 & 4--59 & 36.48 & 5.59 & 817.8 & 319.1 \\
\bottomrule
\end{tabular}
}
\end{table}

\subsection{Data Curation Dynamics}
\label{sec:curation-dynamics}

Our curated \name benchmark contains \datasetsize{} problems, partitioned into
three difficulty tiers based on the number of FSM states.  
Summary statistics are provided in Table~\ref{tab:dataset-stats}.

\paragraph{Generation runtime.}
All semantic FSMs and NL specifications are generated using gpt-5 through
the OpenAI API. The entire generation stage completes quickly under batch
parallelism.  
The dominant computational cost lies in verification: running Yosys’s
equivalence check on a single FSM typically takes
\(\sim\!30\) seconds, making formal checking the primary bottleneck of the pipeline.

\paragraph{Filtering statistics.}
Our pipeline applies two filters: an isomorphism check between the abstract
graph and the LLM-generated YAML, and a equivalence check between the
reference and reconstructed RTL.  
Out of 1{,}500 generated candidates, 1{,}411 (94.1\%) pass the isomorphism test,
and 1{,}085 (76.9\%) also pass RTL equivalence.  
Equivalence-check pass rates decrease with FSM size (95.7\%, 82.1\%, 62.4\% across
the three tiers), but remain sufficiently high to scale further by increasing
generation budget or adopting hierarchical generation for larger FSMs.  
We randomly select 1{,}000 verified examples to form the final \name dataset.

\subsection{Alignment with Real-World Applications}
\label{sec:real-world-alignment}

To validate that \name{} reflects the characteristics of realistic hardware designs, we analyze the distribution of both our NL specifications and the underlying FSM structures.

\paragraph{NL Distribution.}
We demonstrate that the generated specifications align with industrial examples in both descriptive style and length. Qualitatively, our specifications share the narrative structure found in real-world datasheets. 
For instance, we compare the example in Figure~\ref{fig:example} with a standard I2C-Master/Slave Core specification~\cite{i2c-spec}:
\begin{itemize}
    \item \textbf{\name{} Spec:} ``After reset, the controller stays idle until the host asserts a read request (\texttt{rd\_req = 1}), at which point it enters the mode-validation phase.''
    \item \textbf{Real-World I2C Spec:} ``In the idle state, the core leaves the buses free and will be waiting for command. If there is a transaction in MODE bit from '0' to '1', the core will go to start, and will act as Master.''
\end{itemize}
Both descriptions utilize a similar narrative structure. 
Quantitatively, the real-world I2C specification, which describes a 7-state machine, contains 346 words. 
In our dataset, the subset of 7-state FSMs (41 problems) has an average length of 373.1 words. 
This consistency is also observed in other protocols.
These comparisons confirm that our generated specifications closely mirror real-world application standards in terms of verbosity and information density.

\paragraph{FSM Distribution.}
Regarding the structural distribution of the FSMs, we emphasize that \name{} is not merely a static dataset but a fully parameterizable generation framework. 
While the current benchmark utilizes sample-based generation to ensure broad coverage, the underlying framework offers flexible control to address specific distribution requirements:
\begin{itemize}
    \item \textbf{Granular Control over Topology:} The framework allows explicit configuration of state counts, transition densities, and structural constraints. This enables the generation of FSMs that mirror the specific complexity profiles found in target real-world applications.
    \item \textbf{Extensibility to Real-World Seeds:} The pipeline is designed to be extensible; it can ingest FSM structures extracted from real-world legacy code or specifications and use them as seeds to generate diverse, correct-by-construction variations.
\end{itemize}
This flexibility ensures that the tool can adapt to any specific ``real-life'' distribution required, rather than being fixed to a single statistical pattern.

\subsection{Human Check}
\label{sec:human-check}
To further ensure data quality, we perform a manual audit on a subset of examples.  
Each sampled instance is examined along four criteria:
(1) \textit{State Coverage}: the specification must describe every YAML state with no missing or spurious behaviors.
(2) \textit{Transition Coverage}: every YAML transition must be reflected in the specification, with no extra or altered edges.
(3) \textit{Specification-FSM Alignment}: the narrative must allow an unambiguous mapping from each described behavior back to the YAML-specified FSM.
(4) \textit{Hardware Plausibility}: state names, signal names, and contextual descriptions must form a coherent and realistic hardware scenario.
All 20 inspected samples satisfy these criteria, confirming the semantic consistency between the specification and the underlying FSM.

\begin{table*}[t]
\small
\centering
\caption{Model accuracy (\%) across three RTL generation benchmarks:
VerilogEval~v2 \cite{pinckney_2025_revisiting}, RTLLM~v2 \cite{liu_2024_openllmrtl}, and our \name dataset.
For each evaluation pipeline, the best-performing model is shown in \textbf{bold} and the second-best is \underline{underlined}. }
\resizebox{\textwidth}{!}{
\begin{tabular}{
    l|c|c||cccc|cccc|cccc|>{\columncolor{lightgray}}c
}
\toprule
\multirow{2}{*}{\textbf{Model}}&\textbf{Verilog}&
\textbf{RT}
& \multicolumn{4}{c|}{\textbf{Spec $\rightarrow$ RTL}}
& \multicolumn{4}{c|}{\textbf{Spec $\rightarrow$ YAML $\rightarrow$ RTL}} 
& \multicolumn{4}{c|}{\textbf{Spec $\rightarrow$ SystemC}} 
  & \multirow{2}{*}{\cellcolor{white}\textbf{Avg.}}\\
&\textbf{Eval}&\textbf{LLM}& Easy & Med. & Hard & Avg. & Easy & Med. & Hard & Avg. & Easy & Med. & Hard & Avg.\\
\midrule
Llama4-Scout &48.7&36.0&24.0&5.1&0.0&9.7&38.0&4.8&0.3&14.4&0.0&0.0&0.0&0.0&8.0\\
Qwen3-4B &39.1&30.0&20.7&3.6&0.6&8.3&27.5&3.6&0.6&10.6&39.5&11.7&2.1&17.8&12.2\\
Qwen3-8B &53.2&38.0&14.1&5.4&2.7&7.4&44.9&12.9&2.1&20.0&61.4&29.7&5.1&32.1&19.8\\
gpt-5-nano &72.4&54.0&55.4&19.2&4.5&26.4&29.9&5.4&0.6&12.0&58.7&20.4&3.3&27.5&22.0\\
Llama4-Maverick &59.6&50.0&63.2&27.0&6.3&32.2&63.2&24.0&6.6&31.3&35.0&15.0&5.1&18.4&27.3\\
Qwen3-14B &59.0&42.0&55.4&26.4&9.0&30.3&62.0&23.4&4.8&30.1&63.5&24.6&10.2&32.8&31.1\\
gpt-oss-20B &46.8&42.0&61.1&34.8&13.5&36.5&45.8&15.6&3.3&21.6&73.4&37.2&9.3&40.0&32.7\\
gpt-oss-120B &53.8&44.0&77.5&45.6&19.8&47.7&48.5&26.7&8.7&28.0&74.6&44.7&16.8&45.4&40.4\\
Qwen3-32B &64.7&52.0&66.5&37.5&14.4&39.5&77.2&41.1&13.5&44.0&68.9&34.2&12.6&38.6&40.7\\
DeepSeek-R1-0528&72.4&58.0&64.7&41.4&10.8&39.0&68.3&46.5&24.0&46.3&68.0&45.0&4.2&39.1&41.5\\
Gemini-2.5-Flash&60.3&56.0&76.0&55.9&16.5&49.5&86.5&67.0&44.1&65.9&57.8&8.7&0.0&22.2&45.9\\
DeepSeek-V3.1-Terminus&69.9&50.0&80.2&54.7&9.0&48.0&75.4&52.0&24.9&50.8&79.9&46.8&1.8&42.9&47.2\\
Gemini-2.5-Pro&77.6&\underline{60.0}&77.5&59.8&39.6&59.0&87.4&68.2&55.6&70.4&46.1&7.5&0.0&17.9&49.1\\
grok-4-fast-reasoning &74.4&58.0&75.1&48.6&37.8&53.9&78.1&50.5&40.8&56.5&74.0&52.6&39.9&55.5&55.5\\
gpt-5-mini &78.2&54.0&88.6&64.6&41.7&65.0&66.8&36.0&21.6&41.5&\underline{84.4}&60.1&38.7&61.1&55.9\\
Claude-4.5-Haiku &75.0&54.0&84.7&53.2&30.9&56.3&86.5&55.3&33.9&58.6&82.3&54.7&27.9&55.0&56.6\\
gpt-5 &\textbf{86.2}&\textbf{64.0}&\underline{93.1}&\textbf{79.6}&\underline{59.8}&\underline{77.5}&\underline{93.7}&\underline{76.6}&\underline{65.8}&\underline{78.7}&82.0&\underline{73.3}&\textbf{55.0}&\underline{70.1}&\underline{75.4}\\
Claude-4.5-Sonnet &\underline{82.1}&\textbf{64.0}&\textbf{95.5}&\underline{79.3}&\textbf{70.0}&\textbf{81.6}&\textbf{94.0}&\textbf{83.2}&\textbf{72.1}&\textbf{83.1}&\textbf{93.4}&\textbf{80.8}&\underline{54.7}&\textbf{76.3}&\textbf{80.3}\\
\midrule
\textbf{Average}                   &65.2&50.3&65.2&41.2&21.5&42.6&65.2&38.5 & 23.5& 42.3& 63.5&35.9&15.9&38.5&41.1 \\
\bottomrule
\end{tabular}
}

\label{tab:main}
\end{table*}

\subsection{Evaluation Pipeline}
\label{sec:evaluation-pipeline}

We evaluate models in three settings.  
(1) \textbf{Spec → RTL:} the model generates SystemVerilog directly from the NL
specification, and correctness is determined by cycle-accurate agreement with
the reference RTL under the same testbench.  
(2) \textbf{Spec → YAML → RTL:} the model first produces an fsm2sv-compatible
YAML FSM, which is compiled to RTL and evaluated using the same criterion.  
(3) \textbf{Spec → SystemC:} the model outputs a SystemC design, which we test
via SystemC-SystemVerilog co-simulation in Questa; correctness again requires
cycle-by-cycle agreement with the reference RTL.

\section{Evaluation}\label{sec:evaluation}

\subsection{Experimental Setup}

In addition to our \name
benchmark, we also evaluate on two human-written RTL datasets:
VerilogEval~v2~\cite{pinckney_2025_revisiting} and
RTLLM~v2~\cite{liu_2024_openllmrtl}.
We evaluate a broad set of frontier model families, including gpt-5, Claude-4.5, Gemini-2.5, grok-4, Qwen-3, DeepSeek-V3.1/R1, and Llama-4. These models span both proprietary and
open-source families and represent current state-of-the-art LLM systems
across a wide range of model sizes. For all models, we set the maximum output token budget to 16{,}384.
Temperature and top-p follow each model's default settings.
We report Pass@1 as the primary metric, counting a sample as correct only if the generated RTL compiles and passes the reference testbench.

\subsection{Main Results}

\paragraph{\name is a challenging benchmark}
As shown in Figure \ref{tab:main}, across 18 frontier models and three
evaluation pipelines, the overall average Pass@1 is only 41.1\%. Among all
evaluated LLMs, Claude-4.5-Sonnet achieves the highest score of 80.3\%, yet its performance drops to 65.6\% on the hard tier. 
Because our dataset is generated through a fully automated pipeline, increasing the number of states or transitions can produce harder instances, allowing \name to evolve alongside future model improvements.

\paragraph{Different evaluation pipelines lead to sharply different outcomes across models}
For example, Gemini-2.5-Pro attains 70.4\% on the Spec$\rightarrow$YAML$\rightarrow$RTL task but only 17.9\% on the Spec$\rightarrow$SystemC task. Across all evaluated models, the SystemC setting shows the lowest average performance, suggesting that LLMs are less familiar with hardware-compatible high-level languages than with RTL.
By contrast, the Spec$\rightarrow$RTL and Spec$\rightarrow$YAML$\rightarrow$RTL pipelines yield similar average accuracies, indicating that modern LLMs are able to perform finite-state reasoning directly in RTL without explicitly reconstructing the FSM structure in YAML.

\subsection{Analysis}

\begin{figure}[!t]
    \centering
    \begin{subfigure}[t]{0.49\columnwidth}
        \centering
        \includegraphics[width=\linewidth]{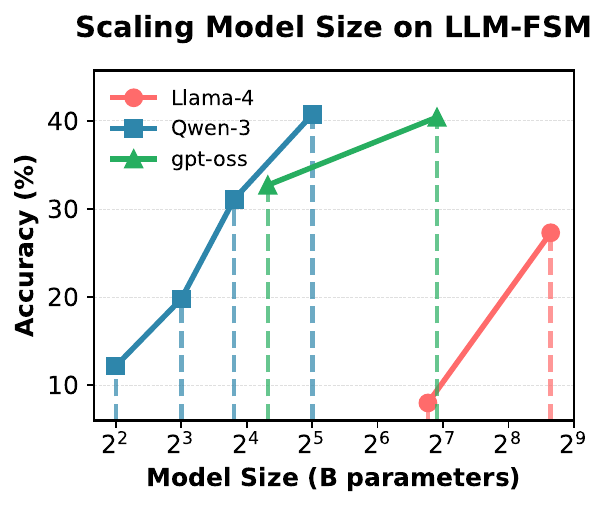}
        \label{fig:scaling_trend_sub1}
    \end{subfigure}\hfill%
    \begin{subfigure}[t]{0.49\columnwidth}
        \centering
        \includegraphics[width=\linewidth]{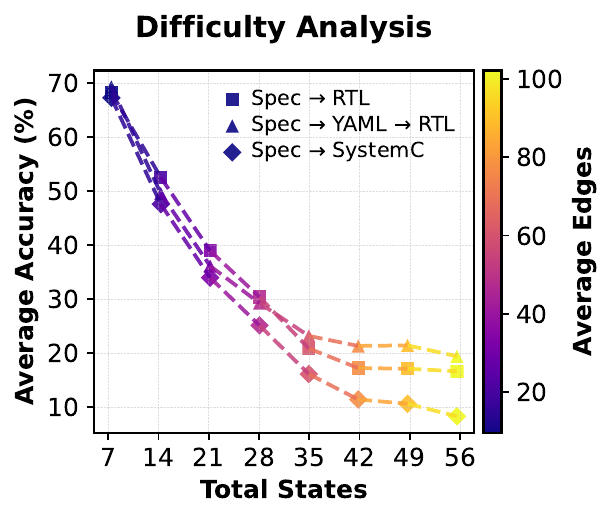}
        \label{fig:scaling_trend_sub2}
    \end{subfigure}
    \vspace{-2em}
    \caption{Scaling and difficulty analysis on \name. Left: scaling behavior of different model families. Right: accuracy averaged across all models within each difficulty bin.}
    \label{fig:scaling_trend}
\end{figure}

\paragraph{Scaling trend and difficulty analysis}
As shown in Figure~\ref{fig:scaling_trend}, within the same family, models benefit from increasing parameter scale. 
On the difficulty axis, accuracy drops sharply as the number of states and edges increases, with the Spec$\rightarrow$SystemC pipeline being the most sensitive.
This confirms that our generation pipeline provides fine-grained control over task difficulty.

\begin{table}[t]
\caption{Correlation between performance on our benchmark and human-written datasets. 
Pipeline~1: Spec$\rightarrow$RTL, Pipeline~2: Spec$\rightarrow$YAML$\rightarrow$RTL, Pipeline~3: Spec$\rightarrow$SystemC.
We report Pearson (P) and Spearman (S) correlations.}
\label{tab:correlation}
\centering
\resizebox{\linewidth}{!}{
\begin{tabular}{l|cc|cc|cc|cc}
\toprule
\multirow{2}{*}{\textbf{Dataset}} &
\multicolumn{2}{c|}{\textbf{Pipeline 1}} &
\multicolumn{2}{c|}{\textbf{Pipeline 2}} &
\multicolumn{2}{c|}{\textbf{Pipeline 3}} &
\multicolumn{2}{c}{\textbf{Overall Avg.}} \\
 & \textbf{P} & \textbf{S} & \textbf{P} & \textbf{S} & \textbf{P} & \textbf{S} & \textbf{P} & \textbf{S} \\
\midrule
VerilogEval & 0.82 & 0.85 & 0.78 & 0.78 & 0.66 & 0.62 & 0.83 & 0.87 \\
RTLLM       & 0.85 & 0.82 & 0.85 & 0.85 & 0.59 & 0.52 & 0.84 & 0.83 \\
\bottomrule 
\end{tabular}}
\vspace{-0.4cm}
\end{table}

\begin{table}[h]
    \centering
    \small
    \setlength{\tabcolsep}{3.5pt}
    \caption{Performance comparison of Base vs. SFT models across difficulty tiers. SFT models were trained only on a subset of Easy/Medium problems, making the Hard tier an OOD test.}
    \label{tab:sft_results}
    \begin{tabular}{l | cc | cc | cc | cc}
        \toprule
        & \multicolumn{2}{c|}{\textbf{Easy (ID)}} & \multicolumn{2}{c|}{\textbf{Medium (ID)}} & \multicolumn{2}{c|}{\textbf{Hard (OOD)}} & \multicolumn{2}{c}{\textbf{Overall}} \\
        \textbf{Model} & \textit{Base} & \textbf{SFT} & \textit{Base} & \textbf{SFT} & \textit{Base} & \textbf{SFT} & \textit{Base} & \textbf{SFT} \\
        \midrule
        Qwen3-4B & 14.9 & \textbf{31.3} & 4.5 & \textbf{10.4} & 0.0 & \textbf{4.5} & 6.5 & \textbf{15.4} \\
        Qwen3-8B & 13.4 & \textbf{14.9} & 6.0 & \textbf{16.4} & 4.5 & \textbf{7.5} & 8.0 & \textbf{12.9} \\
        Qwen3-14B & 55.2 & \textbf{85.1} & 23.9 & \textbf{67.2} & 4.5 & \textbf{34.3} & 27.9 & \textbf{62.2} \\
        \bottomrule
    \end{tabular}
    \vspace{-0.2cm}
\end{table}

\paragraph{Correlation with human-written benchmarks}
To validate that performance on \name reflects real-world RTL generation
ability, we compare model accuracy on our dataset with two human-written
benchmarks: VerilogEval~v2 \cite{pinckney_2025_revisiting} and RTLLM~v2 \cite{liu_2024_openllmrtl}. As shown in Table~\ref{tab:correlation}, model performance exhibits strong positive
correlation across datasets. 
Interestingly, both VerilogEval~v2 and RTLLM~v2 primarily evaluate the direct
Spec$\rightarrow$RTL generation path, and we observe that correlations are
indeed highest for our Pipeline~1 results, further suggesting that \name
captures the same underlying reasoning skills required for hand-written RTL
benchmarks.

\paragraph{Error analysis}
We manually analyze a subset of incorrect generations and identify four
common failure modes.  
(1) \textit{Syntax errors}: models occasionally include invalid syntax in generated RTL code that fails to compile.
(2) \textit{Incorrect timing semantics}: the implementation violates
cycle-level behavior described in the specification, such as transitioning
out of a state earlier or later than required.  
(3) \textit{State or transition mistakes}: the generated code introduces
missing, extra, or reordered edges, resulting in an FSM whose structure
differs from the intended one.
(4) \textit{Formatting errors}: tasks involving YAML or SystemC templates
often fail because the model does not precisely follow the required
schema, leading to invalid FSM descriptions or modules that cannot be
parsed.

\subsection{Training-Time Scaling Results}
\label{sec:training-time-scaling}

\paragraph{Experimental Setup.}
To validate the utility of our generated data for unseen tasks, we conducted an SFT experiment. 
We utilized Qwen3-4B, 8B, and 14B as base models and applied QLoRA fine-tuning. 
For Training Data, we used correct thinking traces generated by Claude-4.5-Sonnet. 
Crucially, we employed an 80/20 split across the entire benchmark. 
For training, we utilized only the 80\% partitions of the \textit{Easy} and \textit{Medium} tiers, while intentionally discarding the training partition of the \textit{Hard} tier to test generalization capabilities on unseen complexity.
We then evaluated the fine-tuned models on the consistent held-out 20\% partitions:
(1) \textit{ID}: the held-out test sets of the \textit{Easy} and \textit{Medium} tiers; and
(2) \textit{OOD}: the held-out test set of the \textit{Hard} tier.

\paragraph{Results and Analysis.}
The results in Table~\ref{tab:sft_results} show consistent improvement across all difficulty levels. 
Specifically, for the Qwen3-4B model, accuracy on the Easy (ID) tier improved from 14.9\% (Base) to 31.3\% (SFT), and on the Medium (ID) tier from 4.5\% to 10.4\%. 
Most importantly, on the unseen Hard (OOD) tier, the model improved from 0.0\% to 4.5\%, raising the Overall accuracy from 6.5\% to 15.4\%. 
Larger models exhibited even stronger generalization, with Qwen3-14B improving from 4.5\% to 34.3\% on the Hard tier.
Furthermore, after fine-tuning on our synthetic data, the Qwen3-4B model was able to correctly answer human-authored FSM-related questions in VerilogEval (e.g., \texttt{Prob100\_fsm3comb}) that the base model previously failed to solve. 
This confirms that \name{} serves as an effective training ground for enhancing performance on real-world, FSM-related hardware tasks.

\subsection{Test-Time Scaling Results}

\begin{figure}[t]
    \centering
    \begin{subfigure}[t]{0.49\columnwidth}
        \centering
        \includegraphics[width=\linewidth]{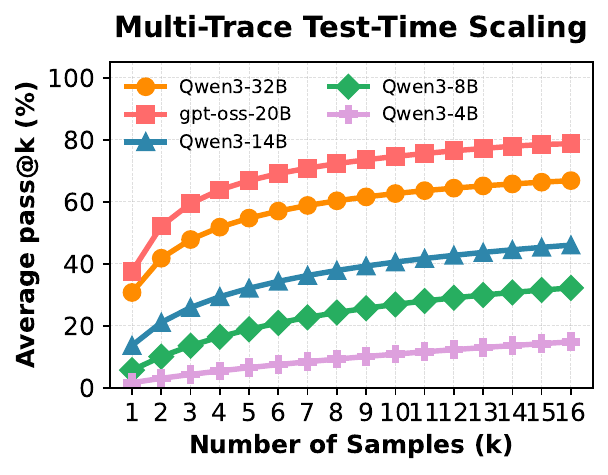}
        \label{fig:scaling_sub1}
    \end{subfigure}\hfill%
    \begin{subfigure}[t]{0.49\columnwidth}
        \centering
        \includegraphics[width=\linewidth]{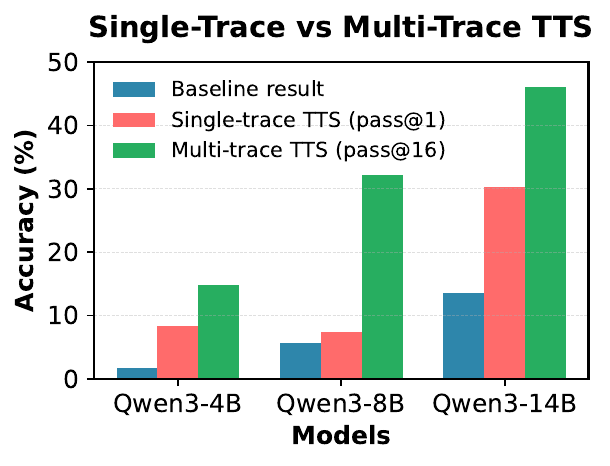}
        \label{fig:scaling_sub2}
    \end{subfigure}
    \vspace{-2em}
    \caption{TTS for finite-state reasoning. Left: multi-trace TTS pass@k scaling on \name. Right: comparison of single-trace TTS vs multi-trace TTS at $k=16$.}
    \label{fig:tts}
    \vspace{-1em}
\end{figure}

As shown in Figure \ref{fig:tts} (left), increasing the number of samples per question in our multi-trace TTS setting steadily improves each model's pass@k on \name. However, simply letting models think before answering (single-trace TTS) is less effective: Figure \ref{fig:tts} (right) shows that Qwen3-14B's pass@1 in thinking mode remains well below its own pass@16 under multi-trace TTS, with smaller models showing the same gap.

\section{Conclusion}\label{sec:conclusion}
In this paper, we introduce \name, a scalable benchmark for generating RTL from NL specifications. 
The dataset explicitly targets finite-state reasoning, provides fine-grained difficulty control, and checks consistency between each specification and its RTL implementation through SAT-based verification. 
Evaluating a wide range of LLMs on \datasetsize{} problems, we find that current models still struggle with temporally precise RTL synthesis. 
Performance on \name{} also correlates with results on human-written RTL benchmarks, indicating that it captures real-world design challenges. 
We further demonstrate that training-time scaling via SFT generalizes effectively to OOD complexity, while multi-trace TTS boosts model performance at inference time.
Overall, \name{} provides a scalable foundation for evaluating and advancing finite-state reasoning in LLM-based RTL generation.
\bibliographystyle{ACM-Reference-Format}
\bibliography{sample-base}

@article{thakur_2023_benchmarking,
  author = {Thakur, Shailja and Ahmad, Baleegh and Fan, Zhenxing and Pearce, Hammond and Tan, Benjamin and Karri, Ramesh and Dolan-Gavitt, Brendan and Garg, Siddharth},
  title = {Benchmarking Large Language Models for Automated Verilog RTL Code Generation},
  year = {2023},
  journal = {DATE}
}

@article{chang_2023_chipgpt,
  author = {Chang, Kaiyan and Wang, Ying and Ren, Haimeng and Wang, Mengdi and Liang, Shengwen and Han, Yinhe and Li, Huawei and Li, Xiaowei},
  title = {ChipGPT: How far are we from natural language hardware design},
  year = {2023},
  journal = {NeurIPS SysML Workshop}
}

@article{thakur_2024_verigen,
  author = {Thakur, Shailja and Ahmad, Baleegh and Pearce, Hammond and Tan, Benjamin and Dolan-Gavitt, Brendan and Karri, Ramesh and Garg, Siddharth},
  title = {VeriGen: A Large Language Model for Verilog Code Generation},
  year = {2024},
  journal = {TODAES}
}

@article{liu_2024_rtlcoder,
  author = {Liu, Shang and Fang, Wenji and Lu, Yao and Zhang, Qijun and Zhang, Hongce and Xie, Zhiyao},
  title = {RTLCoder: Outperforming GPT-3.5 in Design RTL Generation with Our Open-Source Dataset and Lightweight Solution},
  year = {2024},
  journal = {LAD Workshop}
}

@article{pei_2024_betterv,
  author = {Pei, Zehua and Zhen, Hui-Ling and Yuan, Mingxuan and Huang, Yu and Yu, Bei},
  publisher = {Cornell University},
  title = {BetterV: Controlled Verilog Generation with Discriminative Guidance},
  year = {2024},
  journal = {ICML}
}

@article{bardianadimi_2024_a,
  author = {Bardia Nadimi and Zheng, Hao},
  title = {A Multi-Expert Large Language Model Architecture for Verilog Code Generation},
  year = {2024},
  journal = {LAD Workshop}
}

@article{ma_2024_verilogreader,
  author = {Ma, Ruiyang and Yang, Yuxin and Liu, Ziqian and Zhang, Jiaxi and Li, Min and Huang, Junhua and Luo, Guojie},
  title = {VerilogReader: LLM-Aided Hardware Test Generation},
  year = {2024},
  journal = {LAD Workshop}
}

@article{bhandari_2024_llmaided,
  author = {Bhandari, Jitendra and Knechtel, Johann and Narayanaswamy, Ramesh and Garg, Siddharth and Karri, Ramesh},
  title = {LLM-Aided Testbench Generation and Bug Detection for Finite-State Machines},
  year = {2024},
  journal = {Arxiv}
}

@article{qiu_2024_autobench,
  author = {Qiu, Ruidi and Zhang, Grace Li and Drechsler, Rolf and Schlichtmann, Ulf and Li, Bing},
  title = {AutoBench: Automatic Testbench Generation and Evaluation Using LLMs for HDL Design},
  year = {2024},
  journal = {MLCAD}
}

@article{wang_2025_large,
  author = {Wang, Ning and Yao, Bingkun and Zhou, Jie and Wang, Xi and Jiang, Zhe and Guan, Nan},
  title = {Large Language Model for Verilog Generation with Code-Structure-Guided Reinforcement Learning},
  year = {2025},
  journal = {ICLAD}
}

@article{liao_2024_are,
  author = {Liao, Yuchao and Adegbija, Tosiron and Lysecky, Roman},
  publisher = {ACM},
  title = {Are LLMs Any Good for High-Level Synthesis?},
  year = {2024},
  journal = {ICCAD}
}

@article{batten_2024_pyhdleval,
  author = {Batten, Christopher and Pinckney, Nathaniel and Liu, Mingjie and Ren, Haoxing and Brucek Khailany},
  title = {PyHDL-Eval: An LLM Evaluation Framework for Hardware Design Using Python-Embedded DSLs},
  year = {2024},
  journal = {MLCAD}
}

@article{liu_2025_craftrtl,
  author = {Liu, Mingjie and Tsai, Yun-Da and Zhou, Wenfei and Ren, Haoxing},
  publisher = {Cornell University},
  title = {CraftRTL: High-quality Synthetic Data Generation for Verilog Code Models
  with Correct-by-Construction Non-Textual Representations and Targeted Code
  Repair},
  year = {2025},
  journal = {ICLR}
}

@article{qiu_2025_correctbench,
  author = {Qiu, Ruidi and Zhang, Grace and Drechsler, Rolf and Schlichtmann, Ulf and Li, Bing},
  title = {CorrectBench: Automatic Testbench Generation with Functional Self-Correction using LLMs for HDL Design},
  year = {2025},
  journal = {DATE}
}

@article{cui_2024_origen,
  author = {Cui, Fan and Yin, Chenyang and Zhou, Kexing and Xiao, Youwei and Sun, Guangyu and Xu, Qiang and Guo, Qipeng and Liang, Yun and Zhang, Xingcheng and Song, Demin and Lin, Dahua},
  title = {OriGen: Enhancing RTL Code Generation with Code-to-Code Augmentation and Self-Reflection},
  year = {2024},
  journal = {ICCAD}
}

@article{zhao_2025_mage,
  author = {Zhao, Yujie and Zhang, Hejia and Huang, Hanxian and Yu, Zhongming and Zhao, Jishen},
  title = {MAGE: A Multi-Agent Engine for Automated RTL Code Generation},
  year = {2025},
  journal = {DAC}
}

@article{wei_2025_vericoder,
  author = {Wei, Anjiang and Tan, Huanmi and Suresh, Tarun and Mendoza, Daniel and Teixeira, Thiago S. F. X. and Wang, Ke and Trippel, Caroline and Aiken, Alex},
  title = {VeriCoder: Enhancing LLM-Based RTL Code Generation through Functional Correctness Validation},
  year = {2025},
  journal = {NeurIPS DL4C Workshop}
}

@article{wang_2025_verireason,
  author = {Wang, Yiting and Sun, Guoheng and Ye, Wanghao and Qu, Gang and Li, Ang},
  title = {VeriReason: Reinforcement Learning with Testbench Feedback for Reasoning-Enhanced Verilog Generation},
  year = {2025},
  journal = {Arxiv}
}

@article{akyash_2025_rtl,
  author = {Akyash, Mohammad and Azar, Kimia and Kamali, Hadi},
  title = {RTL++: Graph-enhanced LLM for RTL Code Generation},
  year = {2025},
  journal = {ICLAD}
}

@article{deng_2025_scalertl,
  author = {Deng, Chenhui and Tsai, Yun-Da and Liu, Guan-Ting and Yu, Zhongzhi and Ren, Haoxing},
  title = {ScaleRTL: Scaling LLMs with Reasoning Data and Test-Time Compute for Accurate RTL Code Generation},
  year = {2025},
  journal = {MLCAD}
}

@article{yu_2025_spec2rtlagent,
  author = {Yu, Zhongzhi and Liu, Mingjie and Zimmer, Michael and Lin, Yingyan Celine and Liu, Yong and Ren, Haoxing},
  title = {Spec2RTL-Agent: Automated Hardware Code Generation from Complex Specifications Using LLM Agent Systems},
  year = {2025},
  journal = {Arxiv}
}

@article{tasnia_2025_veriopt,
  author = {Tasnia, Kimia and Garcia, Alexander and Farheen, Tasnuva and Rahman, Sazadur},
  title = {VeriOpt: PPA-Aware High-Quality Verilog Generation via Multi-Role LLMs},
  year = {2025},
  journal = {Arxiv}
}

@article{lu_2024_rtllm,
  author = {Lu, Yao and Liu, Shang and Zhang, Qijun and Xie, Zhiyao},
  title = {RTLLM: An Open-Source Benchmark for Design RTL Generation with Large Language Model},
  year = {2024},
  journal = {ASP-DAC}
}

@article{liu_2023_verilogeval,
  author = {Liu, Mingjie and Pinckney, Nathaniel and Khailany, Brucek and Ren, Haoxing},
  title = {VerilogEval: Evaluating Large Language Models for Verilog Code Generation},
  year = {2023},
  journal = {ICCAD}
}

@article{allam_2024_rtlrepo,
  author = {Allam, Ahmed and Shalan, Mohamed},
  title = {RTL-Repo: A Benchmark for Evaluating LLMs on Large-Scale RTL Design Projects},
  year = {2024},
  journal = {LAD Workshop}
}

@article{pinckney_2025_revisiting,
  author = {Pinckney, Nathaniel and Batten, Christopher and Liu, Mingjie and Ren, Haoxing and Khailany, Brucek},
  title = {Revisiting VerilogEval: A Year of Improvements in Large-Language Models for Hardware Code Generation},
  year = {2025},
  journal = {TODAES}
}

@article{liu_2024_openllmrtl,
  author = {Liu, Shang and Lu, Yao and Fang, Wenji and Li, Mengming and Xie, Zhiyao},
  title = {OpenLLM-RTL: Open Dataset and Benchmark for LLM-Aided Design RTL Generation },
  year = {2024},
  journal = {ICCAD}
}

@article{pinckney_2025_comprehensive,
  author = {Pinckney, Nathaniel and Deng, Chenhui and Ho, Chia-Tung and Tsai, Yun-Da and Liu, Mingjie and Zhou, Wenfei and Khailany, Brucek and Ren, Haoxing},
  title = {Comprehensive Verilog Design Problems: A Next-Generation Benchmark Dataset for Evaluating Large Language Models and Agents on RTL Design and Verification},
  year = {2025},
  journal = {Arxiv}
}

@article{purini_2025_archxbench,
  author = {Purini, Suresh and Garg, Siddhant and Gaur, Mudit and Bhat, Sankalp and Mupparapu, Sohan and Ravindran, Arun},
  title = {ArchXBench: A Complex Digital Systems Benchmark Suite for LLM Driven RTL Synthesis},
  year = {2025},
  journal = {MLCAD}
}

@article{calzada_2025_verilogdb,
  author = {Calzada, Paul E. and Ibnat, Zahin and Rahman, Tanvir and Kandula, Kamal and Lu, Danyu and Saha, Sujan Kumar and Farahmandi, Farimah and Tehranipoor, Mark},
  title = {VerilogDB: The Largest, Highest-Quality Dataset with a Preprocessing Framework for LLM-based RTL Generation},
  year = {2025},
  journal = {Arxiv}
}

@article{sun_2023_towards,
  author = {Sun, Chuyue and Hahn, Christopher and Trippel, Caroline},
  title = {Towards Improving Verification Productivity with Circuit-Aware Translation of Natural Language to SystemVerilog Assertions},
  year = {2023},
  journal = {CAV Workshop}
}

@article{yan_2025_assertllm,
  author = {Yan, Zhiyuan and Fang, Wenji and Li, Mengming and Li, Min and Liu, Shang and Xie, Zhiyao and Zhang, Hongce},
  publisher = {ACM},
  title = {AssertLLM: Generating Hardware Verification Assertions from Design Specifications via Multi-LLMs},
  year = {2025},
  journal = {ASP-DAC}
}

@article{cosler_2023_nl2spec,
  author = {Cosler, Matthias and Hahn, Christopher and Mendoza, Daniel and Schmitt, Frederik and Trippel, Caroline},
  publisher = {Springer Science+Business Media},
  title = {nl2spec: Interactively Translating Unstructured Natural Language to Temporal Logics with Large Language Models},
  year = {2023},
  journal = {CAV}
}

@article{mendoza_2024_translating,
  author = {Mendoza, Daniel and Hahn, Christopher and Trippel, Caroline},
  title = {Translating Natural Language to Temporal Logics with Large Language Models and Model Checkers},
  year = {2024},
  journal = {FMCAD}
}

@article{ravindran_2025_survey,
  author = {Ravindran, Arun and Patra, Aditya and Babaey, Vahid and Purini, Suresh},
  title = {Survey and Benchmarking of Large Language Models for RTL Code Generation: Techniques and Open Challenges},
  year = {2025},
  journal = {Preprints}
}

@article{tasnia_2025_opl4gpt,
    author = {Tasnia, Kimia and Rahman, Sazadur},
    title = {OPL4GPT: An Application Space Exploration of Optimal Programming Language for Hardware Design by LLM},
    year = {2025},
  journal = {ASP-DAC}
}

@article{guo2025deepseek,
    author={Guo, Daya
    and Yang, Dejian
    and Zhang, Haowei
    and Song, Junxiao
    and Wang, Peiyi
    and Zhu, Qihao
    and Xu, Runxin
    and Zhang, Ruoyu
    and Ma, Shirong
    and Bi, Xiao
    and Zhang, Xiaokang
    and Yu, Xingkai
    and Wu, Yu
    and Wu, Z. F.
    and Gou, Zhibin
    and Shao, Zhihong
    and Li, Zhuoshu
    and Gao, Ziyi
    and Liu, Aixin
    and ...
    and Zhang, Zhen},
    title={DeepSeek-R1 incentivizes reasoning in LLMs through reinforcement learning},
    journal={Nature},
    year={2025}
}

@article{li_2025_llms,
  author = {Li, Dacheng and Cao, Shiyi and Griggs, Tyler and Liu, Shu and Mo, Xiangxi and Tang, Eric and Hegde, Sumanth and Hakhamaneshi, Kourosh and Patil, Shishir G and Zaharia, Matei and Gonzalez, Joseph E and Stoica, Ion},
  title = {LLMs Can Easily Learn to Reason from Demonstrations Structure, not content, is what matters!},
  year = {2025},
  journal = {Arxiv}
}

@article{wei2022chain,
  title={Chain-of-thought prompting elicits reasoning in large language models},
  author={Wei, Jason and Wang, Xuezhi and Schuurmans, Dale and Bosma, Maarten and Xia, Fei and Chi, Ed and Le, Quoc V and Zhou, Denny and others},
  journal = {NeurIPS},
  year={2022}
}

@article{muennighoff_2025_s1,
  author = {Muennighoff, Niklas and Yang, Zitong and Shi, Weijia and Li, Xiang Lisa and Fei-Fei, Li and Hajishirzi, Hannaneh and Zettlemoyer, Luke and Liang, Percy and Candès, Emmanuel and Hashimoto, Tatsunori},
  title = {s1: Simple test-time scaling},
  year = {2025},
  journal = {EMNLP}
}

@article{brown_2024_large,
  author = {Brown, Bradley and Juravsky, Jordan and Ehrlich, Ryan and Clark, Ronald and Le, Quoc V and Ré, Christopher and Mirhoseini, Azalia},
  title = {Large Language Monkeys: Scaling Inference Compute with Repeated Sampling},
  year = {2024},
  journal = {Arxiv}
}

@article{snell2025scaling,
    title={Scaling LLM Test-Time Compute Optimally Can be More Effective than Scaling Parameters for Reasoning},
    author={Snell, Charlie Victor and Lee, Jaehoon and Xu, Kelvin and Kumar, Aviral},
    journal = {ICLR},
    year={2025},
}

@article{wang2024math,
  title={Math-Shepherd: Verify and Reinforce LLMs Step-by-step without Human Annotations},
  author={Wang, Peiyi and Li, Lei and Shao, Zhihong and Xu, Runxin and Dai, Damai and Li, Yifei and Chen, Deli and Wu, Yu and Sui, Zhifang},
  journal = {ACL},
  year={2024}
}

@article{wangself,
  title={Self-Consistency Improves Chain of Thought Reasoning in Language Models},
  author={Wang, Xuezhi and Wei, Jason and Schuurmans, Dale and Le, Quoc V and Chi, Ed H and Narang, Sharan and Chowdhery, Aakanksha and Zhou, Denny},
  journal = {ICLR},
  year = {2023}
}

@article{wang_2025_ranked,
    title = "Ranked Voting based Self-Consistency of Large Language Models",
    author = "Wang, Weiqin  and
      Wang, Yile  and
      Huang, Hui",
    journal = "ACL Findings",
    year = "2025",
}

@article{yao2023tree,
  title={Tree of thoughts: Deliberate problem solving with large language models},
  author={Yao, Shunyu and Yu, Dian and Zhao, Jeffrey and Shafran, Izhak and Griffiths, Tom and Cao, Yuan and Narasimhan, Karthik},
  journal={NeurIPS},
  year={2023}
}

@article{biforest,
  title={Forest-of-Thought: Scaling Test-Time Compute for Enhancing LLM Reasoning},
  author={Bi, Zhenni and Han, Kai and Liu, Chuanjian and Tang, Yehui and Wang, Yunhe},
  journal={ICML},
  year = {2025},
}

@article{wu2025tom,
  author = {Wu, Yuheng and Xie, Jianwen and Zhang, Denghui and Xu, Zhaozhuo},
  title = {DEL-ToM: Inference-Time Scaling for Theory-of-Mind Reasoning via Dynamic Epistemic Logic},
  year = {2025},
  journal = {EMNLP}
}

@article{wu2025on,
  author = {Wu, Yuheng and Mirhoseini, Azalia and Tambe, Thierry},
  title = {On the Role of Temperature Sampling in Test-Time Scaling},
  year = {2025},
  journal = {Arxiv}
}

@article{fsm2sv,
  author       = {Bamakhrama, Mohamed },
  title        = {fsm2sv: SystemVerilog FSM Generator},
  journal = {GitHub},
  year         = {2021}
}

@article{yosys-eqy,
  author = {Xenia Wolf, Claire},
  title = {Equivalence Checking with Yosys},
  journal = {GitHub},
  year = {2020}
}

@article{i2c-spec,
  author = {TooMuch Semiconductor},
  title = {I2C-Master / Slave Core Specification},
  journal = {OpenCores},
  year = {2007}
}

\end{document}